%% file: acl_latex.tex
\documentclass[11pt]{article}

\usepackage{amsthm, amsmath, amssymb}
\usepackage{mathrsfs}

\usepackage[preprint]{acl}

\usepackage{times}
\usepackage{latexsym}
\usepackage{algorithm}
\usepackage{algorithmicx}
\usepackage{algpseudocode}
\usepackage{booktabs}
\usepackage{multirow}
\usepackage[table]{xcolor}
\usepackage{makecell}
\usepackage[most]{tcolorbox}

\usepackage{tablefootnote}
\usepackage{subcaption}

\usepackage[T1]{fontenc}

\usepackage[utf8]{inputenc}

\usepackage{microtype}

\usepackage{inconsolata}

\usepackage{graphicx}

\title{SteerRM: Debiasing Reward Models via Sparse Autoencoders}

\author{
Mengyuan Sun\thanks{Equal contribution.}, 
Zhuohao Yu\footnotemark[1], 
Weizheng Gu, 
Shikun Zhang, 
Wei Ye\thanks{Corresponding author.} \\
National Engineering Research Center for Software Engineering, Peking University \\
\texttt{\{mengyuansun25, zyu\}@stu.pku.edu.cn},
\texttt{wye@pku.edu.cn}
}

\begin{document}
\maketitle
\begin{abstract}

Reward models (RMs) are critical components of alignment pipelines, yet they exhibit biases toward superficial stylistic cues, preferring better-presented responses over semantically superior ones. 
Existing debiasing methods typically require retraining or architectural modifications, while direct activation suppression degrades performance due to representation entanglement.
We propose SteerRM, the first training-free method for debiasing reward models using Sparse Autoencoder (SAE)-based interventions.
SteerRM isolates stylistic effects using contrastive paired responses, identifies bias-related SAE features with a strength–stability criterion, and suppresses them at inference time. 
Across six reward models on RM-Bench, SteerRM improves Hard-split accuracy by 7.3 points on average while preserving overall performance. Results on a Gemma-based reward model and a controlled non-format bias further suggest generalization across RM architectures and bias types. 
We further find that format-related features are concentrated in shallow layers and transfer across models, revealing shared architecture-level bias encoding patterns. 
These results show that SAE-based interventions can mitigate reward-model biases without retraining, providing a practical and interpretable solution for alignment pipelines.
\footnote{Our code is available at \url{https://anonymous.4open.science/r/SteerRM}}

\end{abstract}

\input{sec1-intro}
\input{sec2-related}
\input{sec3-methodology}
\input{sec4-experiments}
\input{sec5-conclusion}

\bibliography{custom}

\newpage

\input{appendix}

\end{document}

%% file: sec1-intro.tex
\section{Introduction}

Reward models (RMs) are a foundational component of modern alignment pipelines such as reinforcement learning from human feedback (RLHF)~\citep{ouyang2022training, christiano2017deep, stiennon2020learning, ziegler2019fine}. 
By serving as learned proxies for human preferences, RMs guide policy optimization and implicitly define what behaviors are reinforced during training~\citep{zhong2025comprehensive,yu2025reward, wang2024secrets}.  
Despite their central role, recent evidence shows that RMs are not purely semantic evaluators: they exhibit systematic preferences for superficial attributes of responses, including length, verbosity, and formatting~\citep{lambert2025rewardbench, malik2025rewardbench, liu2025rmbench, casper2023open}. 
\emph{Format bias} refers to the phenomenon where RMs assign systematically different scores to responses that share \emph{identical semantic content} but differ only in \emph{surface formatting}, such as Markdown versus plain text~\citep{liu2025rmbench}. This bias manifests when RMs assign higher scores to factually incorrect but well-formatted responses than to correct but plainly formatted answers. Format bias distorts preference signals and incentivizes models to optimize presentation over correctness~\citep{chen2024odin, taylor2025school}.

Existing approaches to mitigating RM bias predominantly operate through training-time modifications~\citep{dubois2024length, bu2025beyond}, architectural changes~\citep{shen-etal-2023-loose}, or post hoc score calibration~\citep{huang2025posthoc, park2024offsetbias}. While effective in some cases, these methods treat the RM as a black box, often requiring retraining or additional supervision.
Direct internal interventions, like suppressing activation differences, are challenging due to highly entangled representations.
Sparse Autoencoders (SAEs) provide a promising alternative by decomposing representations into sparse, interpretable features that enable targeted interventions without retraining~\citep{bricken2023monosemantic, lieberum2024gemma, rajamanoharan2024jumping}. 
Prior work has applied SAE-based steering to large language models for behavior control~\citep{templeton2024scaling, chalnev2024improving, shu2025survey, bhalla2024towards}, showing that SAE features capture semantically meaningful directions and enable precise behavior modification. However, the use of pretrained SAE dictionaries for mitigating non-semantic biases such as formatting or stylistic preferences in reward models has received little attention.

In this work, we introduce \textbf{SteerRM}, a training-free method for mitigating format bias in reward models using SAE-based interventions. Our key insight is that format preferences are often concentrated in a small subset of SAE features that respond reliably to systematic cues. This perspective reframes reward model debiasing as a representation-editing problem: by identifying and selectively suppressing bias-related SAE features, we can directly steer the reward model’s preferences away from undesirable directions while preserving its ability to evaluate semantic content, all without modifying model parameters or training objectives. While SteerRM is developed around Markdown formatting, an additional controlled study suggests that the same pipeline can \textbf{extend to other stylistic confounders}.

SteerRM implements this through a systematic \textbf{three-stage pipeline}. First, we synthesize format-controlled paired responses to isolate formatting effects. Second, we identify format-sensitive SAE features using a strength-stability criterion that selects features consistently activating on formatting cues. Third, we steer the reward model by suppressing these features during inference, effectively neutralizing format preference without updating model parameters.

This work makes three key contributions. 
\textbf{First}, we propose the first training-free method for debiasing reward models using SAE-based interventions, overcoming the representation entanglement limitations of direct activation suppression that cause severe performance degradation. 
\textbf{Second}, through SAE-based analysis, we demonstrate that format-related SAE features are localized in shallow Transformer layers and are transferable across different models, indicating they encode shared surface-level signals.
\textbf{Third}, SteerRM improves RM-Bench Hard split accuracy by 7.3 points on average across six LLaMA-based reward models while maintaining overall stability, with additional tests on a Gemma-based model and a different stylistic confounder supporting generalization across architectures and bias types.

%% file: sec2-related.tex
\section{Related Work}

\begin{figure*}[t]
  \centering
  \includegraphics[width=\textwidth]{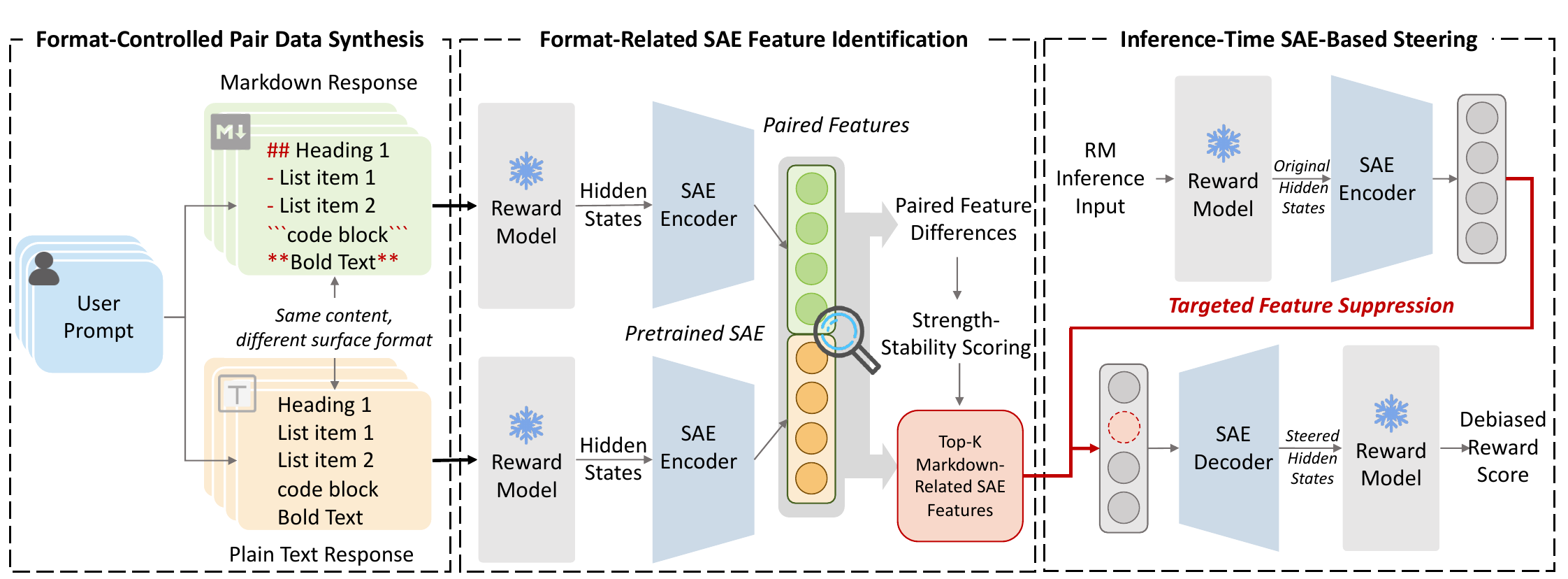}
  \caption{Overview of SteerRM. Our training-free pipeline consists of three stages:
  (1) synthesizing paired responses with different surface formats or styles, (2) identifying format-related SAE
  features, and (3) suppressing these features at inference time.}
  \label{fig:pipeline}
\end{figure*}

\subsection{Reward Models}

Reward models (RMs) serve as learned proxies for human preferences in modern alignment pipelines~\citep{ouyang2022training,stiennon2020learning}, typically finetuned to assess prompt–response pairs via scalar reward heads~\citep{liu2024skywork,dorka2024quantile} or generative objectives~\citep{wang2024pandalm,kim2023prometheus,yu2025rewardanything}.
However, trained RMs are susceptible to spurious correlations from superficial attributes such as length, verbosity, politeness, and formatting~\citep{gao2023scaling, bai2022training, casper2023open}. RM-Bench~\citep{liu2025rmbench} shows that under style-controlled Hard settings, many RMs exhibit near-random accuracy, indicating reliance on stylistic cues rather than semantic quality.


Prior work on mitigating superficial biases in reward models largely falls into two categories.
One line of work addresses bias through training-time or architectural interventions, such as explicit length control~\citep{dubois2024length}, mixture-of-experts designs to disentangle stylistic preferences from content quality~\citep{shen-etal-2023-loose}, or dynamic weighting of length as a context-dependent factor~\citep{bu2025beyond}.
A second line focuses on post hoc calibration, correcting reward scores after inference to reduce the influence of surface features. Key methods include statistical correction based on regression analysis~\citep{huang2025posthoc} and leveraging debiased datasets to tune evaluator weights as seen in OffsetBias~\citep{park2024offsetbias}.

However, these methods often rely on additional training or architectural changes, increasing computational cost and complexity. By comparison, representation-level interventions that directly manipulate internal activations at inference time remain underexplored.

\subsection{Sparse Autoencoders}

Sparse Autoencoders (SAEs) are pre-trained interpretability tools that decompose LLM activations into sparse, human-understandable features~\citep{bricken2023monosemantic}. An SAE processes hidden states $\mathbf{h}_t$ into sparse feature vectors $\mathbf{f}_t$ through a training objective that combines reconstruction loss with $L_1$ sparsity regularization:
\begin{equation}
\mathcal{L} = \underbrace{\|\mathbf{h}_t - \text{Dec}(\mathbf{f}_t)\|^2}_{\mathcal{L}_{\text{rec}}} + \lambda \underbrace{\|\mathbf{f}_t\|_1}_{\mathcal{L}_{\text{sparse}}},
\end{equation}
where $\lambda$ controls the sparsity trade-off. This training produces interpretable features~\citep{bricken2023monosemantic, lieberum2024gemma}. Pretrained SAEs are now widely available~\citep{lieberum2024gemma, he2024llama} and generalize effectively to instruction-tuned models~\citep{kissane2024saes, templeton2024scaling,he2024llama}, enabling feature-level analysis without additional training.

Several studies have explored leveraging SAE features to modify language model behavior, for example by steering along decoder directions~\citep{templeton2024scaling}, manipulating specific activation patterns~\citep{chalnev2024improving}, or using correlation-based feature selection for generation-time steering~\citep{cho2025corrsteer}. 
Extending SAEs to reward modeling, SARM~\citep{zhang2025interpretable} and SparseRM~\citep{liu2025sparserm} leverage sparse features to build new reward models, but both require retraining or architectural modifications and focus on model construction rather than debiasing existing reward models. 

Although pretrained SAE feature dictionaries are widely available, their use for mitigating biases such as formatting or stylistic preferences in reward models remains unexplored.

%% file: sec3-methodology.tex
\section{Methodology}
\label{sec:method}

\subsection{Problem Setup and Overview}

We consider a setting in which a reward model (RM) assigns different scores to responses that share the same semantic content but differ in formatting. Let $f_\theta(x,y)\in\mathbb{R}$ denote a fixed RM scoring a response $y$ to a prompt $x$. For each prompt $x$, consider a format-controlled pair $(y^{md}, y^{pl})$ that is matched in content and differs only in surface form (Markdown versus plain text). The format-induced score gap is defined as
\begin{equation}
    \Delta(x) = f_\theta(x,y^{md}) - f_\theta(x,y^{pl}),
\end{equation}
and our objective is to reduce such gaps without updating the RM parameters $\theta$.

SteerRM is a training-free \textbf{analysis-and-intervention} pipeline that operates directly on the RM's internal hidden representations using pretrained open-source SAE dictionaries. As illustrated in Figure~\ref{fig:pipeline}, the framework consists of three stages: (1) \textit{data synthesis} to generate content-matched response pairs, (2) \textit{feature identification} to localize bias-relevant SAE features, and (3) \textit{feature intervention} to suppress these signals during inference. While this work focuses on Markdown bias, the pipeline is inherently generalizable and can be extended to mitigate other superficial artifacts such as verbosity, politeness, or specific stylistic cues by synthesizing appropriate paired data to isolate the corresponding SAE features.

\subsection{Format-Controlled Pair Data Synthesis}
\label{sec:method:data}

To isolate formatting as the sole varying factor, we synthesize a paired dataset $\mathcal{D}=\{(x_i, y_i^{md}, y_i^{pl})\}_{i=1}^N$, where each triple consists of a user prompt $x_i$, a Markdown-formatted response $y_i^{md}$, and a plain-text counterpart $y_i^{pl}$ that preserves the lexical content of $y_i^{md}$ with Markdown markup removed.
This design enables paired comparisons in which any change in RM behavior can be attributed to formatting rather than content variation.

We generate paired responses using a large language model with prompts that explicitly enforce content matching. The model is instructed to first produce a Markdown response and then derive a plain-text version by removing only Markdown syntax, without paraphrasing or adding content. We synthesize data across multiple domains, including chat, reasoning, math, and code, to avoid domain-specific artifacts. Full prompt templates are provided in Appendix~\ref{app:prompts}.

To ensure data quality and diversity, we first deduplicated prompts based on cosine similarity of their sentence embeddings\footnote{Using \texttt{all-MiniLM-L6-v2}~\citep{reimers2019sentence}}, filtering out highly similar queries. We then applied strict regex validation to verify that $y_i^{md}$ contained valid Markdown syntax while $y_i^{pl}$ was completely free of formatting markers. Finally, a manual audit of 50 random pairs confirmed that the plain-text versions preserved the original information content without semantic drift.

\subsection{Format-Related SAE Feature Identification}
\label{sec:method:identify}

Given the paired dataset, we identify SAE features associated with Markdown formatting.
We first extract hidden representations from the RM and encode them into SAE latents. Let $h_\ell(x,y)\in\mathbb{R}^{T\times d}$ denote the hidden activation sequence at Transformer layer $\ell$ when scoring the concatenated prompt-response text $(x,y)$.
We extract $h_\ell$ via forward hooks on the Transformer blocks.

For each target layer $\ell$, we load pretrained Sparse Autoencoders, each consisting of an encoder $E_\ell$ and decoder $D_\ell$. Given token-wise hidden states, we compute token-wise SAE latents
\begin{equation}
z_\ell(x,y) = E_\ell(h_\ell(x,y))\in\mathbb{R}^{T\times m}.
\end{equation}
These latents are aggregated into a single feature vector by averaging over non-special tokens. Special tokens such as BOS, EOS, and PAD are excluded because they serve structural purposes rather than encoding content-related information, and their activations would introduce noise into format-related feature analysis. Let $M\in\{0,1\}^{T}$ denote a token mask where $M_t=1$ for non-special tokens and $M_t=0$ for special tokens:
\begin{equation} 
\bar z_\ell(x,y) = \frac{1}{|\mathcal{T}|}\sum_{t\in\mathcal{T}} z_{\ell,t}(x,y)\in\mathbb{R}^{m}, 
\end{equation}
where $\mathcal{T} = \{t : M_t=1\}$ denotes the set of non-special token positions. The vectors $\bar z_\ell$ are concatenated across layers to obtain $\bar z(x,y)$.

For each paired example $(x_i, y_i^{md}, y_i^{pl})$, we compute a paired difference for every SAE feature:
\begin{equation}
d_{i,j} = \bar z_j(x_i, y_i^{md}) - \bar z_j(x_i, y_i^{pl}),
\end{equation}
where $j$ indexes SAE features. A positive $d_{i,j}$ indicates stronger activation under Markdown formatting for matched content.

We score features using a \textbf{strength-stability} criterion. Strength is measured by the mean paired difference $\mu_j = \mathbb{E}_i[d_{i,j}]$, while stability is measured by the variance $\sigma_j^2 = \mathrm{Var}_i[d_{i,j}]$ across the dataset. We normalize these quantities globally across all layers to $[0,1]$ using min-max normalization to preserve cross-layer comparability, and define the feature score as
\begin{equation}
\text{score}_j = \bar{\mu}_j \cdot (\bar{\sigma}_j + \epsilon)^{-1},
\end{equation}
where $\epsilon$ is a small constant for numerical stability. A global top-$K$ selection is then performed across all layers, retaining features with $\mu_j>0$, which are more strongly associated with Markdown formatting. The resulting layer-wise feature sets $\{\mathcal{S}_\ell\}$ are used for downstream steering interventions.

\begin{figure*}[t]
  \centering
  \includegraphics[width=2\columnwidth]{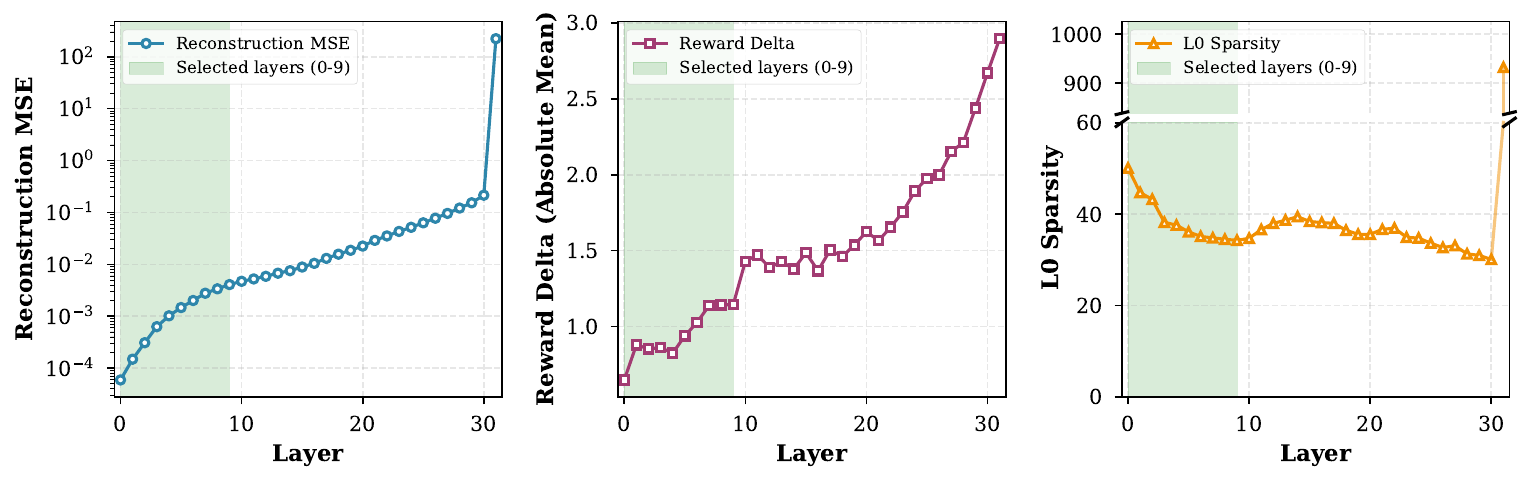}
  \caption{SAE reconstruction quality across layers. 
(Left) Reconstruction MSE. 
(Middle) Absolute reward difference between original and reconstructed scores. 
(Right) L0 sparsity measured as the mean number of active features per sample. Layers 0-9 are selected based on these metrics.
}
  \label{fig:layer_selection}
\end{figure*}

\subsection{Reward Model Steering via SAE}
\label{sec:method:intervene}

At inference time, SteerRM performs feature-level steering by ablating selected SAE latents and reconstructing the corresponding hidden representations. Selected features are zeroed in the latent space, and the modified latents are decoded to yield edited hidden states. This reconstruction-based intervention keeps representations consistent with the SAE's learned representation space, while selectively removing format-related signals. The selected features are the Markdown-formatting-related features identified in Section~\ref{sec:method:identify}. Ablating these features steers RM representations toward the plain-text direction and reduces format bias in scoring.

Steering is implemented via forward hooks registered on Transformer blocks at selected layers $\ell$. When a hidden sequence $h_\ell\in\mathbb{R}^{T\times d}$ is produced, the hook intercepts it and applies the intervention: SAE latents $z_\ell = E_\ell(h_\ell)$ are computed, the identified feature coordinates $\mathcal{S}_\ell$ are zeroed out, and the modified latents are decoded to yield an edited reconstruction $\tilde h_\ell = D_\ell(z_\ell^{(0)})$, where
\begin{equation}
(z_\ell^{(0)})_{t,j} =
\begin{cases}
0, & j\in\mathcal{S}_\ell \\
(z_\ell)_{t,j}, & \text{otherwise.}
\end{cases}
\end{equation}
The original hidden state is then replaced with this reconstruction: $h_\ell' = \tilde h_\ell$.
By reconstructing from modified latents, the edited representation remains within the SAE's learned manifold, maintaining coherence while removing format-related feature contributions.

To maintain consistency with feature identification, the intervention is applied only on non-special-token positions using the same token mask $M\in\{0,1\}^{T}$ as defined in Section~\ref{sec:method:identify}:
\begin{equation}
h_{\ell,t}' =
\begin{cases}
\tilde h_{\ell,t}, & M_t=1\\
h_{\ell,t}, & M_t=0.
\end{cases}
\end{equation}

All interventions are training-free: RM parameters remain fixed, and SteerRM reuses pretrained SAE dictionaries for deterministic activation editing at inference.

%% file: sec4-experiments.tex
\section{Experiments}

Our experiments investigate: 
(1) Where are format features \emph{localized} in reward models? (2) Can SteerRM \emph{reduce format bias} without \emph{compromising performance}? (3) Is SAE decomposition \emph{necessary} for effective debiasing? (4) Do format features \emph{transfer} across models?

\paragraph{Experimental Setup}

We evaluate SteerRM on six reward models: Skywork-Reward-Llama-3.1-8B~\citep{liu2024skywork}, QRM-Llama3.1-8B-v2~\citep{dorka2024quantile}, URM-LLaMa-3.1-8B~\citep{lou2024uncertainty}, Llama-3.1-8B-Base-RM-RB2~\citep{malik2025rewardbench}, Llama-3.1-8B-Instruct-RM-RB2~\citep{malik2025rewardbench}, and Llama-3.1-Tulu-3-8B-RM~\citep{lambert2024tulu}. 
These RMs are based on either Llama-3.1-8B or Llama-3.1-8B-Instruct~\citep{dubey2024llama}, and all are sequence-classification models that output scalar scores for prompt-response pairs.

For SAE-based steering, we use pretrained LlamaScope Sparse Autoencoders~\citep{he2024llama} corresponding to the Llama-3.1-8B base architecture. These SAEs are trained on the base model's activations and provide layer-wise feature dictionaries for representation decomposition.

Our main study centers on the Llama-3.1 family because open-source pretrained SAEs are readily available for this backbone and many recent open reward models are built on it. To test cross-architecture generalization, we additionally evaluate SteerRM on a Gemma-based reward model using Gemma Scope~\citep{lieberum2024gemma}; details are in Appendix~\ref{app:gemma_setup}.

Following the pipeline described in Section~\ref{sec:method}, we first synthesize 500 format-controlled response response pairs using GPT-4.1 mini~\citep{openai2025gpt41}. Each pair contains a prompt with both Markdown and plain-text responses that preserve identical semantic content. This probing set is used to identify format-related SAE features.

We apply the strength–stability scoring criterion and perform global top-$K$ selection with $K=10$, balancing feature coverage with intervention precision. Robustness to probing set size, data source, and the choice of $K$ is analyzed in Appendices~\ref{app:probe_size}, \ref{app:existing_samples}, and \ref{app:topk_ablation}.

We evaluate all models on RM-Bench~\citep{liu2025rmbench}, a benchmark designed to assess reward model sensitivity to subtle changes and robustness to style bias. RM-Bench includes four domains (Chat, Math, Code, Safety) and three difficulty levels (Easy, Normal, Hard). For each prompt, the benchmark provides chosen and rejected responses with varying styles, enabling evaluation of both preference accuracy and format bias.

\subsection{Layer Selection and Feature Localization}
\label{sec:exp:layer}

Selecting appropriate Transformer layers is a key design choice in SteerRM.
Because the SAEs are pretrained on Llama-3.1-8B, we evaluate their ability to reconstruct reward model activations, which may differ from those of the base model.

We assess SAE reconstruction quality by measuring three key metrics across all layers: reconstruction error (MSE), reward score preservation (reward delta), and L0 sparsity (the number of active SAE features). Our measurement settings for MSE and L0 sparsity align with established practices in the SAE literature~\citep{he2024llama, kissane2024saes}. Detailed evaluation settings are provided in Appendix~\ref{app:eval_settings}. 

Figure~\ref{fig:layer_selection} summarizes these metrics across layers.
Layers 0–9 consistently exhibit good reconstruction quality: reconstruction MSE (log scale) ranges from $5.95 \times 10^{-5}$ to $4.09 \times 10^{-3}$, reward delta stays relatively stable between $0.65$ and $1.14$, and L0 sparsity remains moderate at $34.3$ to $50.1$ active features per sample.
In contrast, all three metrics degrade sharply beyond layer 9, with reconstruction error increasing by orders of magnitude, reward delta becoming highly variable, and L0 sparsity showing erratic spikes.
These results suggest a clear representation-level distinction. Lower layers (0–9) preserve token- and structure-level representations that are largely shared between the base model and the reward model, enabling reliable SAE reconstruction. In contrast, higher layers encode increasingly model-specific patterns that diverge from base model representations, likely reflecting the effects of reward-model fine-tuning.

\begin{figure}[t]
  \centering
  \includegraphics[width=.9\columnwidth]{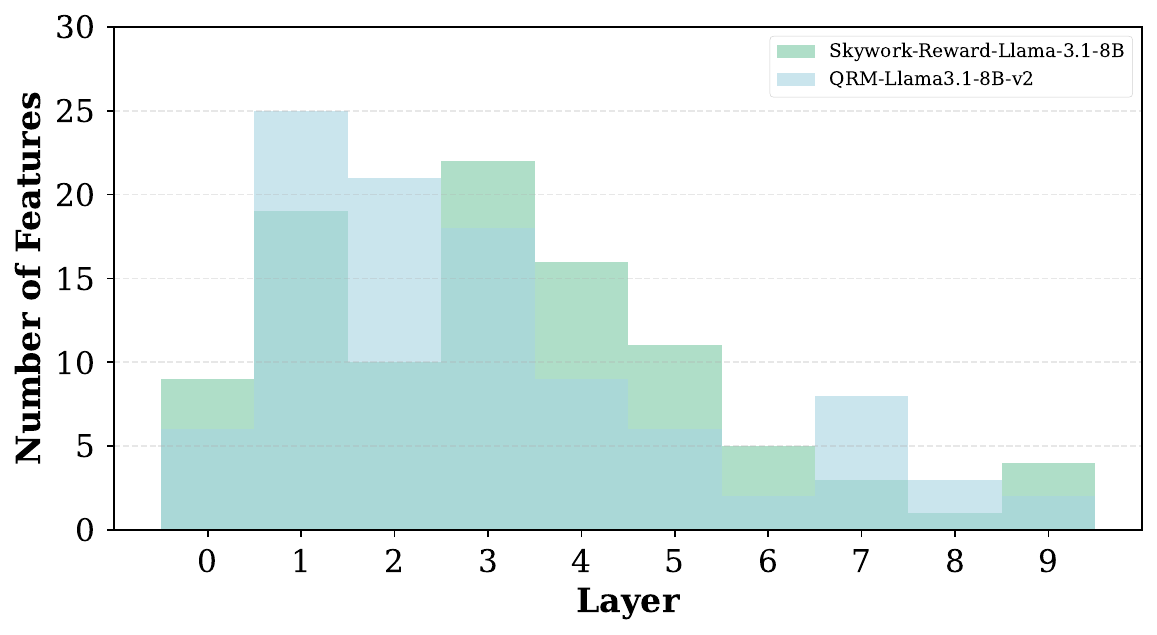}
  \caption{Distribution of top-100 candidate format-related SAE features across layers. Format-related features are concentrated in early layers (1-3), revealing that formatting information is encoded at shallow layers of the Transformer.}
  \label{fig:feature_distribution}
\end{figure}

We therefore restrict feature identification to layers 0-9. Within this range, we apply the feature identification procedure described in Section~\ref{sec:method:identify} and perform global top-$K$ selection with $K=10$ across all candidate layers. 
To understand the layer-wise distribution of format-related features, we analyze the top-100 candidate features across layers 0-9. Figure~\ref{fig:feature_distribution} shows this distribution for two models, which exhibit consistent patterns across all evaluated models. The results reveal that format-related features are concentrated in the early layers, with the majority of top-ranked candidates located in layers 1-3. 
This localization aligns with the hierarchical organization of Transformer representations~\citep{jawahar2019does, tenney2019bert}. Formatting cues, such as Markdown markers, are surface-level signals typically encoded in early layers. While these initial layers focus on local syntactic patterns, higher layers progressively integrate them into abstract semantic representations where format and content become increasingly intertwined.

\paragraph{\textbf{Takeaway 1.}} Format-related features are primarily concentrated within the initial layers of the Transformer (Layers 1–3), revealing that format biases are encoded as surface-level signals in the shallow representation hierarchy.

\input{tables/main_res}

\subsection{Main Results}
\label{sec:exp:main}

Table~\ref{tab:rmbench_main} presents the main results of SteerRM across six reward models on RM-Bench. SteerRM consistently improves Hard split accuracy across all models (gains of 3.4 to 10.3 points), where format-enhanced incorrect responses are paired with format-plain correct ones. This improvement demonstrates that suppressing format-related features eliminates format-based differences, allowing models to evaluate responses based solely on semantic content.

Easy split accuracy drops by 5.1 points on average, reflecting the removal of a format-based shortcut in which correct responses typically exhibit better formatting, leading to inflated baseline performance. In contrast, Normal split performance remains largely stable with an average change of 1.0 point. Because responses in this split share identical formatting, it provides a format-fair evaluation that isolates content quality. The observed stability confirms that SteerRM suppresses format-related bias without degrading the reward model’s core content evaluation, indicating an effective disentanglement of format and content signals.

Across domains, Chat (+0.8), Math (+0.3), and Code (+1.3) show modest average gains. These domain scores represent averages across Easy, Normal, and Hard splits, where substantial Hard improvements are balanced by the overall performance distribution. Safety performance remains largely unaffected (0.8 point change), reflecting that safety judgments rely on strong content signals that naturally dominate format-related noise.

\paragraph{Generalization Across Architectures.} Although our main study focuses on the Llama family, the method only requires a compatible pretrained SAE dictionary rather than architecture-specific retraining. We therefore apply SteerRM to \texttt{Ray2333/GRM-Gemma2-2B-sftreg}~\citep{yang2024regularizing}, a Gemma-based reward model, using Gemma Scope SAEs~\citep{lieberum2024gemma}; full settings are given in Appendix~\ref{app:gemma_setup}. Table~\ref{tab:gemma_main} shows that SteerRM improves Hard accuracy from 39.1\% to 44.8\% while keeping Normal performance essentially unchanged and slightly improving the overall average, indicating that the method is not restricted to Llama-based reward models.

\input{tables/gemma}

\paragraph{Generalization Beyond Formatting.} We additionally evaluate SteerRM on a controlled politeness-bias setting, with construction details deferred to Appendix~\ref{app:politeness_setup}. As shown in Table~\ref{tab:politeness_bias}, adding minor politeness markers to rejected responses reduces Skywork accuracy from 65.5\% to 61.0\%, while SteerRM recovers it to 66.0\%, suggesting that the same contrastive SAE intervention can mitigate controlled stylistic biases beyond Markdown formatting.

\input{tables/politeness}

\paragraph{\textbf{Takeaway 2.}} SteerRM reduces format bias while preserving the reward model's general content evaluation, and additional results on a Gemma-based reward model and another controlled stylistic confounder suggest that the framework generalizes across both model architectures and bias types.

\subsection{Ablation: Necessity of SAE Decomposition}
\label{sec:exp:comparison}


A natural alternative to SAE-based intervention is to directly suppress activation differences between biased and unbiased responses. This approach first computes a bias direction vector for each target layer by averaging the difference between markdown and plain activations across the same paired format-controlled dataset used for SAE feature identification: $\mathbf{d}_\ell = \mathbb{E}[\mathbf{h}_\ell^{md} - \mathbf{h}_\ell^{pl}]$, where $\mathbf{h}_\ell^{md}$ and $\mathbf{h}_\ell^{pl}$ denote hidden states for markdown and plain responses respectively. During inference, this direction is subtracted from the hidden states at each layer: $\mathbf{h}_\ell' = \mathbf{h}_\ell - \mathbf{d}_\ell$. While conceptually straightforward, this direct activation suppression suffers from fundamental limitations due to representation entanglement.

Table~\ref{tab:rmbench_main} reveals a critical pattern: while activation suppression achieves substantial improvements on the Hard split, it simultaneously causes severe degradation on Normal and Easy splits, leading to overall average accuracy dropping to near-random levels (47.6\%$\sim$59.5\%) across all six models. This asymmetric performance pattern stems from the entangled nature of neural representations. The mean difference vector $\mathbf{d}_\ell$ inevitably includes components correlated with both format and content quality, and subtracting this direction indiscriminately suppresses not only format bias but also semantic signals necessary for content evaluation.

In contrast, SAE-based feature suppression uses sparse decomposition to disentangle representations, enabling targeted intervention on format-related features while largely preserving semantic evaluation. Moreover, SAE features are inherently interpretable, with each feature corresponding to a distinct direction in representation space, allowing us to analyze their alignment with specific Markdown syntax, as shown in the next section.

\paragraph{\textbf{Takeaway 3.}} Sparse feature decomposition is a prerequisite for effective debiasing. Without it, direct activation suppression suffers from representation entanglement, leading to catastrophic performance collapse and near-random accuracy.

\subsection{Feature Transferability Analysis}
\label{sec:exp:transferability}

\input{tables/trans}

We evaluate the transferability of format-related SAE features across reward models sharing the same base architecture. Specifically, we identify markdown-related SAE features on Skywork-Reward-Llama-3.1-8B and apply the same feature set to five other models during intervention. 

As shown in Table~\ref{tab:transferability}, features identified on Skywork consistently transfer to all five target models, yielding substantial gains on the Hard split (e.g., +14.2\% on URM$^5$ and +28.4\% on Tulu$^8$). This cross-model effectiveness indicates that format bias arises from shared architecture-level representations rather than model-specific training artifacts.

Comparing Table~\ref{tab:transferability} with Table~\ref{tab:rmbench_main}, we observe a larger drop in Easy-split accuracy and the Average score under cross-model transfer. In both settings, Easy-split degradation is partly caused by removing stylistic shortcuts. The stronger decline during transfer arises from differences in how reward models utilize these shared format features. Although markdown-related representations are common across the Llama-3.1-8B family, their contributions to reward scoring vary by model. As a result, features identified on Skywork effectively suppress format bias on the Hard split but may also remove signals that target models such as Tulu use to assess clarity or structure, leading to larger drops on Easy and lower Average scores.

To verify that the transferred features capture genuine formatting signals, we analyze their Neuronpedia~\citep{neuronpedia} interpretations and find that top-ranked features consistently activate on markdown syntax elements such as code block delimiters, list markers, and heading indicators.

\begin{tcolorbox}[
    colback=cyan!2!white,
    colframe=cyan!25!gray,
    title={\textbf{Example: Feature 10175 (LLAMASCOPE-RES-131K Layer 3) -- Markdown Bold Markers}},
    fonttitle=\bfseries\small,
    left=6pt,
    right=6pt,
    top=6pt,
    bottom=6pt,
    boxrule=0.5pt,
    breakable
]
\textbf{Semantic Interpretation from GPT-4o-mini:} sections highlighting key features or attributes of items or concepts

\textbf{Top Activation Examples:}
\begin{itemize}
    \item ``we'll talk with two influential Cash App team members: \colorbox{green!30}{**}Liang Shi\colorbox{green!30}{**} on the Marketing team and \colorbox{green!30}{**}Rebecca Corcillo\colorbox{green!30}{**}''
    \item ``Modern Home with great curb appeal and layout, with many recent upgrades! \colorbox{green!30}{**}Fresh new paint throughout\colorbox{green!30}{**} Updated dream kitchen with gas stove''
    \item ``bathrooms hardwood floors in bedrooms, new hardwood floors in living room, \colorbox{green!30}{**}New Central Heat and Air\colorbox{green!30}{**}, Large Detached two car garage!''
\end{itemize}

\emph{The consistent activation on ** markers across diverse contexts confirms this feature's role in detecting markdown bold formatting.}
\end{tcolorbox}

Additional examples in Appendix~\ref{app:neuronpedia_features} further confirm that our criterion isolates format-related representations.

\paragraph{\textbf{Takeaway 4.}} Format-related SAE features exhibit strong cross-model transfer, suggesting that formatting cues are encoded as stable early-layer representations shared across models. This allows reuse of a fixed feature set for bias mitigation, with remaining performance differences driven by model-specific reliance on these representations.

%% file: tables/main_res.tex
\begin{table*}[t]
\centering
\small
\caption{
\textbf{Main results on RM-Bench.}
We compare baseline reward models, direct activation suppression (Activation), and SAE-based feature intervention (SteerRM).
SteerRM consistently improves performance on the Hard split while preserving Normal and Average accuracy, whereas Activation improves Hard performance at the cost of substantial degradation on other splits.
}
\label{tab:rmbench_main}
\setlength{\tabcolsep}{6pt}
\begin{tabular}{l l cccc|cc>{\columncolor{gray!10}}c|c}
\toprule
\textbf{Model} & \textbf{Method} & Chat & Math & Code & Safety & Easy & Normal & Hard & Average \\
\midrule

\multirow{3}{*}{Skywork\tablefootnote{Skywork = \texttt{Skywork/Skywork-Reward-Llama-3.1-8B}.}}
& Baseline   & 69.8 & 60.6 & 54.5 & 96.5 & 88.9 & 74.9 & 47.3 & 70.3 \\
& Activation & 58.0 & 50.3 & 50.0 & 50.1 & 20.8 & 54.0 & 81.5 & 52.1 \\
& SteerRM    & 72.5 & 62.5 & 56.0 & 95.0 & 83.9 & 74.1 & 56.5 & 71.5 \\
\midrule

\multirow{3}{*}{QRM\tablefootnote{QRM = \texttt{nicolinho/QRM-Llama3.1-8B-v2}.}}
& Baseline   & 67.4 & 63.2 & 52.0 & 95.6 & 87.8 & 73.4 & 47.5 & 69.5 \\
& Activation & 47.9 & 49.4 & 47.2 & 46.0 & 46.8 & 46.4 & 49.6 & 47.6 \\
& SteerRM    & 65.5 & 62.9 & 52.8 & 95.2 & 81.1 & 71.2 & 54.9 & 69.1 \\
\midrule

\multirow{3}{*}{URM\tablefootnote{URM = \texttt{LxzGordon/URM-LLaMa-3.1-8B}.}}
& Baseline   & 72.2 & 61.6 & 53.4 & 94.7 & 83.9 & 73.7 & 53.7 & 70.5 \\
& Activation & 58.1 & 49.7 & 47.7 & 51.8 & 20.3 & 54.2 & 80.9 & 51.8 \\
& SteerRM    & 72.9 & 62.0 & 54.6 & 93.6 & 78.3 & 73.0 & 61.0 & 70.8 \\
\midrule

\multirow{3}{*}{Base-RM\tablefootnote{Base-RM = \texttt{allenai/Llama-3.1-8B-Base-RM-RB2}.}}
& Baseline   & 71.0 & 59.6 & 58.3 & 90.8 & 85.5 & 73.3 & 51.0 & 69.9 \\
& Activation & 44.4 & 49.4 & 47.8 & 55.3 & 43.1 & 48.7 & 55.7 & 49.2 \\
& SteerRM    & 71.3 & 59.3 & 56.5 & 89.6 & 81.3 & 71.8 & 54.4 & 69.2 \\
\midrule

\multirow{3}{*}{Ins-RM\tablefootnote{Ins-RM = \texttt{allenai/Llama-3.1-8B-Instruct-RM-RB2}.}}
& Baseline   & 66.8 & 65.0 & 57.0 & 91.6 & 90.7 & 75.8 & 43.8 & 70.1 \\
& Activation & 55.8 & 51.5 & 46.2 & 84.7 & 65.1 & 60.9 & 52.6 & 59.5 \\
& SteerRM    & 67.1 & 65.1 & 59.9 & 91.1 & 87.1 & 75.1 & 50.1 & 70.8 \\
\midrule

\multirow{3}{*}{Tulu\tablefootnote{Tulu = \texttt{allenai/Llama-3.1-Tulu-3-8B-RM}.}}
& Baseline   & 64.8 & 60.2 & 57.7 & 83.5 & 91.9 & 73.8 & 33.9 & 66.5 \\
& Activation & 52.5 & 50.1 & 49.1 & 44.3 & 53.9 & 48.6 & 44.4 & 49.0 \\
& SteerRM    & 67.7 & 60.4 & 60.8 & 83.4 & 86.4 & 73.6 & 44.2 & 68.1 \\

\bottomrule
\end{tabular}
\end{table*}

%% file: tables/gemma.tex
\begin{table}[t]
\centering
\small
\caption{
\textbf{Cross-architecture evaluation on a Gemma-based reward model.}
We apply SteerRM to \texttt{Ray2333/GRM-Gemma2-2B-sftreg} using Gemma Scope SAEs. \textit{Delta} denotes SteerRM minus Baseline.
}
\label{tab:gemma_main}
\setlength{\tabcolsep}{7pt}
\begin{tabular}{l c c c c}
\toprule
\textbf{Method} & \textbf{Easy} & \textbf{Normal} & \textbf{Hard} & \textbf{Average} \\
\midrule
Baseline & 87.6 & 68.2 & 39.1 & 65.0 \\
SteerRM & 83.3 & 68.1 & 44.8 & 65.4 \\
\textit{Delta} & -4.3 & -0.1 & \textbf{+5.7} & \textbf{+0.4} \\
\bottomrule
\end{tabular}
\end{table}

%% file: tables/politeness.tex
\begin{table}[t]
\centering
\small
\caption{
\textbf{Generalization to politeness bias on Skywork.}
\textit{Politeness-Injected} denotes the adversarial variant with minor politeness markers added to rejected responses. Relative change is computed against the baseline accuracy on the original clean set.
}
\label{tab:politeness_bias}
\setlength{\tabcolsep}{5pt}
\begin{tabular}{l l c c}
\toprule
\textbf{Method} & \textbf{Test Set} & \textbf{Accuracy} & \textbf{Rel. Change} \\
\midrule
Baseline & Original & 65.5 & -- \\
Baseline & Politeness-Injected & 61.0 & -6.9\% \\
SteerRM & Politeness-Injected & 66.0 & +0.8\% \\
\bottomrule
\end{tabular}
\end{table}

%% file: tables/trans.tex
\begin{table}[t]
    \centering
    \footnotesize
    \caption{
    \textbf{Feature transferability across models.}
    Format-related SAE features identified on Skywork$^3$ are applied to other reward models.
    We report Baseline and SteerRM accuracies on the Easy, Normal, and Hard splits, as well as the Average score.
    SteerRM consistently improves performance on the Hard split across target models, indicating effective cross-model transfer of format-related features.
    }
    \label{tab:transferability}
    \setlength{\tabcolsep}{3.2pt}
    \renewcommand{\arraystretch}{1.15}

    \resizebox{\columnwidth}{!}{%
    \begin{tabular}{
        >{\centering\arraybackslash}m{1.3cm}
        >{\centering\arraybackslash}m{1.15cm}
        >{\centering\arraybackslash}m{0.75cm}
        >{\centering\arraybackslash}m{0.85cm}
        >{\centering\arraybackslash}m{0.75cm}
        >{\centering\arraybackslash}m{0.95cm}
    }
    \toprule
    \textbf{Target} & \textbf{Method}
    & \textbf{Easy}
    & \textbf{Normal}
    & \textbf{Hard}
    & \textbf{Average} \\
    \midrule

    \multirow{2}{*}{QRM\footnotemark[4]}
    & Baseline & 87.8 & 73.4 & 47.5 & 69.5 \\
    & SteerRM  & 75.9 & 70.0 & 56.1 & 67.3 \\
    \midrule

    \multirow{2}{*}{URM\footnotemark[5]}
    & Baseline & 83.9 & 73.7 & 53.7 & 70.5 \\
    & SteerRM  & 72.7 & 72.6 & 67.9 & 71.1 \\
    \midrule

    \multirow{2}{*}{Base-RM\footnotemark[6]}
    & Baseline & 85.5 & 73.3 & 51.0 & 69.9 \\
    & SteerRM  & 72.4 & 70.1 & 52.3 & 64.9 \\
    \midrule

    \multirow{2}{*}{Ins-RM\footnotemark[7]}
    & Baseline & 90.7 & 75.8 & 43.8 & 70.1 \\
    & SteerRM  & 75.9 & 75.7 & 52.0 & 67.9 \\
    \midrule

    \multirow{2}{*}{Tulu\footnotemark[8]}
    & Baseline & 91.9 & 73.8 & 33.9 & 66.5 \\
    & SteerRM  & 54.2 & 63.4 & 62.3 & 60.0 \\
    \bottomrule
    \end{tabular}%
    }
\end{table}

%% file: sec5-conclusion.tex
\section{Conclusion}

In this work, we study format bias in reward models from a representation-level perspective and show that it can be mitigated through training-free SAE interventions. We find that formatting cues are concentrated in shallow, transferable features, enabling SteerRM to suppress them without modifying reward-model parameters or training objectives. Across the main Llama-based evaluation, SteerRM improves robustness on format-confounded comparisons while preserving general content evaluation, and additional results on a Gemma-based reward model and a controlled stylistic confounder suggest broader generalization. The method directly reuses open-source pretrained SAEs, providing a practical and interpretable complement to retraining-based debiasing.

%% file: appendix.tex
\appendix

\section{Data Synthesis}

\subsection{Prompt Templates}
\label{app:prompts}
To construct the paired dataset, we employed a robust prompt design consisting of a system instruction and domain-specific user prompts. We present the full prompts below.

\paragraph{System Prompt.} The system prompt defines the task of generating paired Markdown and plain-text answers in JSON format.

\begin{quote}
\small\ttfamily
You are a data synthesis assistant responsible for generating training data.

Task (two steps):\\
1) Generate a unique question (prompt) and an answer written in Markdown (answer\_markdown).\\
2) Remove Markdown formatting from answer\_markdown to generate a plain-text version (answer\_plain).

Requirements:\\
- Use Markdown naturally for readability. Start the answer with a normal sentence (do not begin with a heading); you may use headings later if helpful.\\
- answer\_plain must preserve the exact wording/content of answer\_markdown.\\
- Only remove Markdown syntax/markup tokens. Do NOT rewrite, paraphrase, add, or delete content.

Output format as JSON:\\
\{\\
\hspace*{1em}"prompt": "...",\\
\hspace*{1em}"answer\_markdown": "...",\\
\hspace*{1em}"answer\_plain": "..."\\
\}
\end{quote}

\paragraph{Domain-Specific Instructions.} To ensure diversity, we use specific instructions for four domains: \textbf{Code}, \textbf{Math}, \textbf{Reasoning}, and \textbf{Chat}.

\begin{itemize}
    \item \textbf{Code}: \texttt{Generate a unique programming-related question (such as algorithm implementation, code explanation, debugging, etc.). Be creative with the specific topic, programming language, and difficulty level.}
    
    \item \textbf{Math}: \texttt{Generate a unique mathematics question (such as geometry, algebra, calculus, etc.). Be creative with the specific topic and difficulty level.}

    \item \textbf{Reasoning}: \texttt{Generate a unique reasoning question (such as logical reasoning, causal analysis, etc.). Be creative with the specific topic.}
    
    \item \textbf{Chat}: \texttt{Generate a unique general conversation question (such as life advice, knowledge Q\&A, etc.). Be creative with the specific topic.} 
\end{itemize}

\paragraph{User Prompt.} The final user prompt combines the domain instruction with a reinforcement of the requirements.

\begin{quote}
\small\ttfamily
[Domain-specific instruction] 

Generate ONE complete training sample. Requirements recap:\\
- prompt: clear, specific, unambiguous.\\
- answer\_markdown: a natural assistant answer written in Markdown.\\
- answer\_plain: plain text version of answer\_markdown with ALL Markdown removed; keep the exact same wording/content.\\
- Do not add/omit content between the two.

Output only JSON, no other content.
\end{quote}

\subsection{Dataset Statistics}

Starting with an initial pool of 1,000 synthesized pairs (250 per domain), we applied the filtering pipeline detailed in Section~\ref{sec:method:data}. Since the number of valid samples varied slightly across domains after filtering, we downsampled each category to exactly 125 samples. This yielded a final balanced dataset of 500 pairs across the four domains (Code, Math, Reasoning, Chat). A representative example from the final dataset is shown below.

\begin{tcolorbox}[
    colback=gray!5!white,
    colframe=gray!50!black,
    title={\textbf{Example Synthesized Data Pair (JSON)}},
    fonttitle=\bfseries\small,
    left=6pt,
    right=6pt,
    top=6pt,
    bottom=6pt,
    boxrule=0.5pt,
    breakable
]
\small\ttfamily
\{\\
\hspace*{1em}"prompt": "I'm planning a surprise birthday party for my friend. What are some tips to ensure everything goes smoothly?",\\
\\
\hspace*{1em}"answer\_markdown": "That sounds like a fun event! Here are some tips:$\backslash$n$\backslash$n\#\# Planning Tips$\backslash$n$\backslash$n- **Choose a date and time** that works for most guests.$\backslash$n- **Send out invitations early**.$\backslash$n- **Select a suitable venue** based on the number of guests.$\backslash$n$\backslash$nRemember to have fun and focus on creating a memorable experience!",\\
\\
\hspace*{1em}"answer\_plain": "That sounds like a fun event! Here are some tips:$\backslash$n$\backslash$n Choose a date and time that works for most guests.$\backslash$n Send out invitations early.$\backslash$n Select a suitable venue based on the number of guests.$\backslash$n$\backslash$nRemember to have fun and focus on creating a memorable experience!"\\
\}
\end{tcolorbox}

\section{SAE Details and Generalization Analysis}

\subsection{Pretrained SAE Architecture}

We utilize the pretrained Sparse Autoencoders provided by \textit{Llama Scope}~\citep{he2024llama}, which are trained on the hidden representations of the Llama-3.1-8B-Base model. This suite consists of 256 SAEs employing an improved Top-K architecture, covering all 32 Transformer layers. For each layer, models are trained at multiple locations, including the \textbf{Post-MLP Residual Stream} and the \textbf{Feed-Forward Network (MLP)}. Furthermore, the suite offers two dictionary sizes for multi-scale analysis: an $8\times$ expansion (32k features) and a $32\times$ expansion (131k features) relative to the model's hidden dimension ($d=4096$).

In this work, we adopt the \textbf{32$\times$ expansion factor} (131,072 features) to ensure a high-resolution decomposition of the reward model's internal representations. To determine the optimal training location, we evaluate between the Post-MLP Residual Stream and the Feed-Forward Network (MLP). A generalization analysis in the following section assesses which SAE variant better preserves the reward model’s original behavior during reconstruction.

\subsection{SAE Generalization Evaluation}
\label{app:eval_settings}

\begin{figure*}[t]
  \centering
  \includegraphics[width=1.0\textwidth]{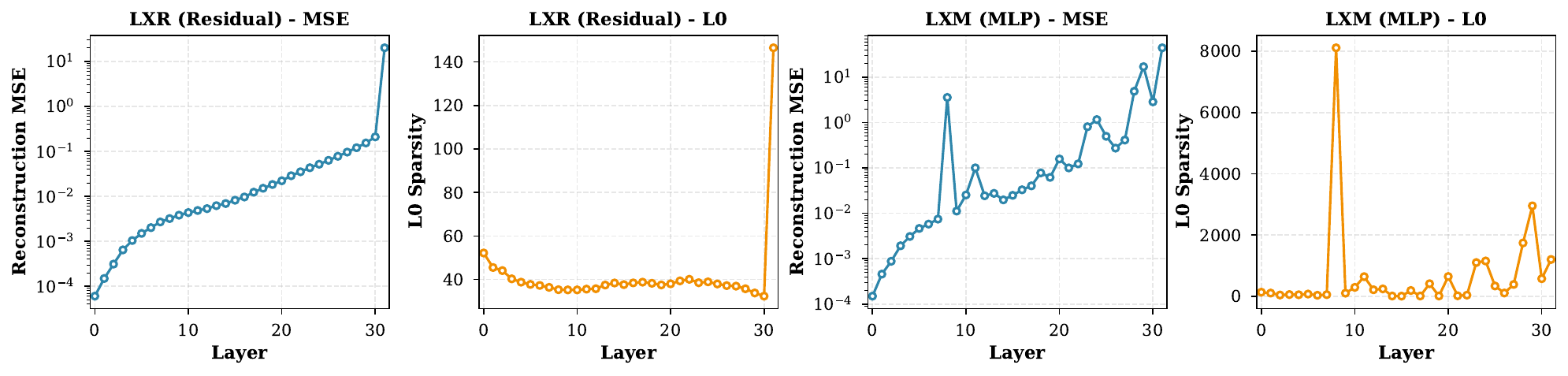}
  \caption{Comparison of SAE reconstruction metrics on the \textbf{Base Model} (Llama-3.1-8B-Base). While both variants achieve comparable MSE, the Residual Stream SAEs (LXR) demonstrate more stable and lower L0 sparsity compared to the MLP Output SAEs (LXM), indicating a more efficient sparse representation.}
  \label{fig:base_metrics}
\end{figure*}

\begin{figure*}[t]
  \centering
  \includegraphics[width=1.7\columnwidth]{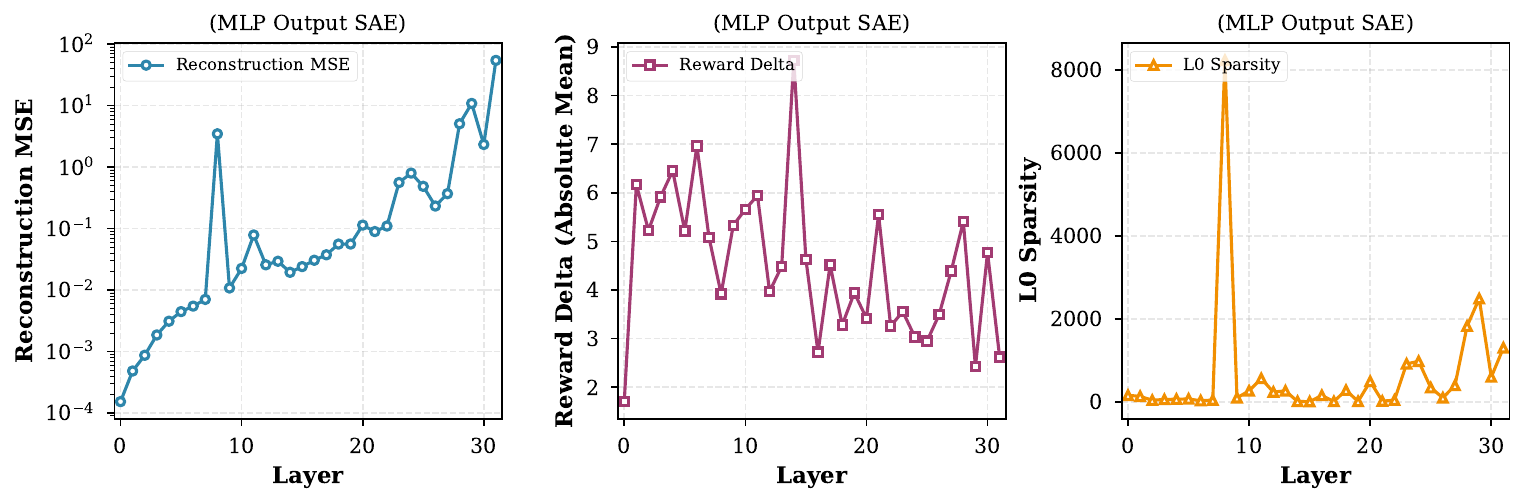}
  \caption{Reconstruction quality for SAEs trained on the \textbf{MLP output (LXM)} when evaluated on the \textbf{Reward Model}. Compared to the Residual Stream SAEs (Figure~\ref{fig:layer_selection}), these models show higher instability in reward preservation (Reward Delta) and sparsity patterns across layers.}
  \label{fig:lxm_metrics}
\end{figure*}

\paragraph{\textbf{Settings.}} To assess the generalization capability of the pretrained SAEs, we conduct a comparative evaluation on both the original Llama-3.1-8B-Base model and the Skywork-Reward model using real-world queries from the \textbf{WildChat} dataset~\citep{zhao2024wildchat}. This dual-evaluation setup allows us to disentangle the SAE's intrinsic reconstruction quality from its transfer performance on the reward model. All SAE-related experiments are conducted using the SAE Lens~\citep{bloom2024saetrainingcodebase}.
\begin{itemize}
    \item \textbf{Base Model Evaluation}: We first evaluate SAEs on the base model to establish a performance baseline. Metrics include \textbf{Reconstruction MSE} (mean squared error between original and reconstructed hidden states) and \textbf{L0 Sparsity} (average number of active features per token)~\citep{bricken2023monosemantic,karvonen2025saebench}.
    \item \textbf{Reward Model Evaluation}: We then evaluate the same SAEs on the reward model to measure transfer robustness. In addition to MSE and L0 sparsity, we compute the \textbf{Reward Score Consistency (Reward Delta)}, defined as the absolute difference between the reward score derived from original representations and that from SAE-reconstructed representations: $\Delta r = |f(h) - f(\hat{h})|$. This metric is crucial for ensuring that the SAE preserves the specific reward modeling function despite the distribution shift from the base model.
\end{itemize}

\paragraph{\textbf{Results.}}
First, we assess the transferability of the SAEs by directly comparing their performance on the base model versus the reward model. We observe that for the Residual Stream (LXR) SAEs, the reconstruction MSE and L0 sparsity metrics on the reward model (Figure~\ref{fig:layer_selection}) closely match those on the base model (Figure~\ref{fig:base_metrics}, left panels). This strong alignment confirms that the features learned from the base model effectively generalize to the reward model, capturing the shared representational structure.

Next, we compare the two SAE variants. While both LXR and LXM variants achieve low reconstruction error on the base model, the LXR SAEs demonstrate significantly better stability. On the base model, LXR variants maintain lower L0 sparsity compared to LXM, indicating a more efficient decomposition. Crucially, when transferred to the reward model, the LXR SAEs exhibit minimal reward score deviation (Reward Delta), whereas the LXM SAEs (Figure~\ref{fig:lxm_metrics}) show higher instability and erratic sparsity patterns.

Given the validated transferability and superior functional preservation, we select the \textbf{Post-MLP Residual Stream SAEs (LXR, 32x expansion)} for all subsequent analyses.

\section{Reward Model Details}

Our main evaluation uses six reward models based on the Llama-3.1-8B architecture, with with an additional Gemma-based model included included for cross-architecture evaluation. Together, they cover both base- and instruction-initialized backbones and a second architecture family. The models are:

\begin{itemize}
    \item \textbf{Skywork/Skywork-Reward-Llama-3.1-8B}~\citep{liu2024skywork}: A high-performance reward model initialized from the Instruct checkpoint. It is trained on the curated Skywork Reward Data Collection (containing 80k high-quality samples sourced from HelpSteer2~\citep{wang2024helpsteer2preferencecomplementingratingspreferences}, Magpie~\citep{xu2024magpie}, and WildGuard~\citep{wildguard2024}), achieving top-tier performance on the RewardBench leaderboard.
    \item \textbf{nicolinho/QRM-Llama3.1-8B-v2}~\citep{dorka2024quantile}: A distributional reward model based on Quantile Regression. It uses Skywork-Reward-Llama-3.1-8B-v0.2~\citep{liu2024skywork} as its backbone and is trained to model reward distributions by aggregating attribute scores, offering a distributional perspective on reward modeling.
    \item \textbf{LxzGordon/URM-LLaMa-3.1-8B}~\citep{lou2024uncertainty}: An uncertainty-aware reward model fine-tuned from Skywork-Reward-Llama-3.1-8B. It employs a two-stage training process: first learning uncertainty-aware attribute distributions on HelpSteer2, and then optimizing a gating layer on the Skywork-Reward-Preference-80K~\citep{liu2024skywork} dataset to aggregate five specific attributes for the final score.
\item \textbf{allenai/Llama-3.1-8B-Base-RM-RB2}~\citep{malik2025rewardbench}: A standard classifier reward model released with RewardBench 2~\citep{malik2025rewardbench}, trained on binary preference data directly on top of the Llama-3.1-8B-Base architecture. This model serves as a baseline for reward modeling without instruction-tuning priors, optimized for correlating with downstream RLHF performance.
    \item \textbf{allenai/Llama-3.1-8B-Instruct-RM-RB2}~\citep{malik2025rewardbench}: A reward model from the RewardBench 2 suite, trained on top of the instruction-tuned Llama-3.1-8B-Instruct. It leverages instruction-following priors to enhance reward modeling capabilities and serves as a counterpart to the base-initialized RM.
    \item \textbf{allenai/Llama-3.1-Tulu-3-8B-RM}~\citep{lambert2024tulu}: The reward model component of the open-source Tulu 3~\citep{lambert2024tulu} alignment suite. It is fine-tuned from the Llama-3.1-Tulu-3-8B-SFT~\citep{lambert2024tulu} checkpoint using a mix of public, synthetic, and human-created preference datasets, representing a modern post-training pipeline.
    \item \textbf{Ray2333/GRM-Gemma2-2B-sftreg}~\citep{yang2024regularizing}: A Gemma-based reward model from the Generalizable Reward Model line, which improves reward-model generalization by regularizing hidden states during training. This checkpoint is fine-tuned from \texttt{gemma-2-2b-it}.
\end{itemize}

The Llama-based suite ensures that our main findings on format feature localization are not artifacts of a single training pipeline, while the additional Gemma-based model provides a complementary cross-architecture check.

\section{Additional Experimental Results}

\subsection{Cross-Architecture Evaluation on Gemma}
\label{app:gemma_setup}

To assess whether SteerRM depends on the Llama-3.1 backbone used in the main study, we additionally evaluate it on \texttt{Ray2333/GRM-Gemma2-2B-sftreg}~\citep{yang2024regularizing}, a reward model built on Gemma 2 2B, using the open-source Gemma Scope SAEs~\citep{lieberum2024gemma}. This setting allows us to test the same intervention pipeline with a different architecture while still using a pretrained public SAE suite.

We follow the same overall procedure as in the main experiments: synthesize 500 format-controlled pairs, rank SAE features with the strength--stability criterion, select a global top-$K$ feature set with $K=10$, and intervene at inference time without updating model parameters. The summary results are reported in Table~\ref{tab:gemma_main} in Section~\ref{sec:exp:main}.

\subsection{Controlled Evaluation Beyond Formatting}
\label{app:politeness_setup}

To test whether SteerRM extends beyond Markdown formatting, we construct a controlled politeness-bias evaluation on the Math and Code subsets of RM-Bench~\citep{liu2025rmbench}, where response quality is anchored to objective correctness. We first filter these subsets to plain-text-only response pairs to remove formatting as a confounder. This yields 757 candidate instances, from which we randomly sample 200 for evaluation.

We use Skywork-Reward-Llama-3.1-8B as the test model. To quantify politeness bias, we build an adversarial variant with GPT-4.1 mini~\citep{openai2025gpt41}, which inserts minor politeness markers such as ``glad to help'' into the rejected response while preserving its incorrect semantic content. These edits are intentionally lightweight so that the underlying correctness of the pair remains unchanged.

For intervention, we apply the same SteerRM pipeline under politeness control. We synthesize response pairs that differ only in politeness level, identify politeness-related SAE features with the same strength-stability criterion used in the main experiments, and suppress these features at inference time. The corresponding results are reported in Table~\ref{tab:politeness_bias} in Section~\ref{sec:exp:main}.

\subsection{Distribution of Format-Related SAE Features}

\begin{figure}[t]
    \centering
    \begin{subfigure}[b]{0.48\columnwidth}
        \centering
        \includegraphics[width=\linewidth]{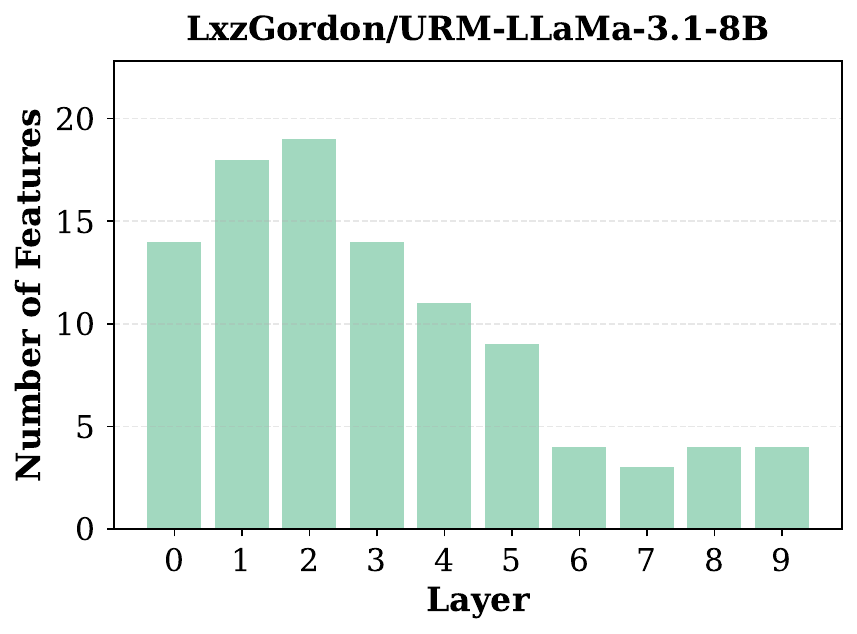}
        \caption{URM-LLaMa-3.1-8B}
        \label{fig:urm}
    \end{subfigure}
    \hfill
    \begin{subfigure}[b]{0.48\columnwidth}
        \centering
        \includegraphics[width=\linewidth]{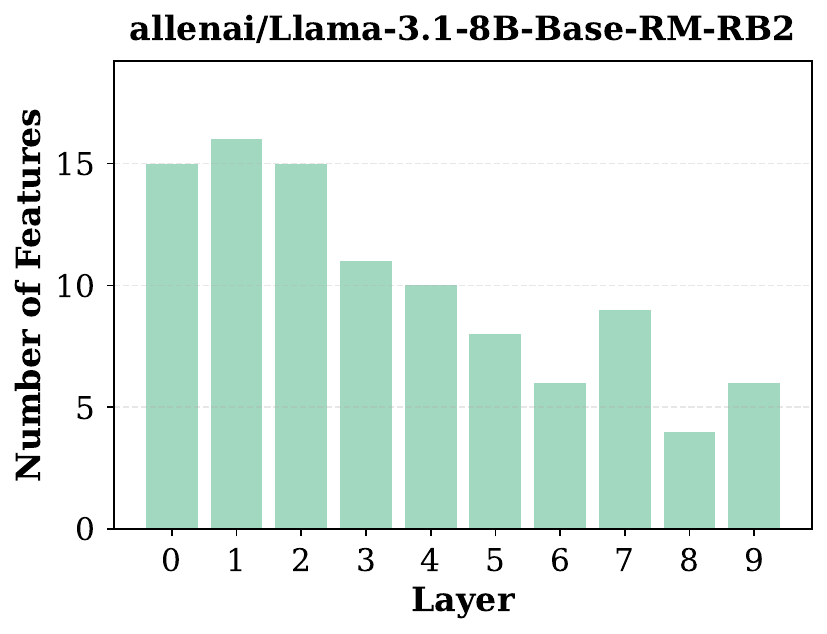}
        \caption{Llama-3.1-8B-Base}
        \label{fig:base}
    \end{subfigure}
    
    \vspace{0.5em}
    
    \begin{subfigure}[b]{0.48\columnwidth}
        \centering
        \includegraphics[width=\linewidth]{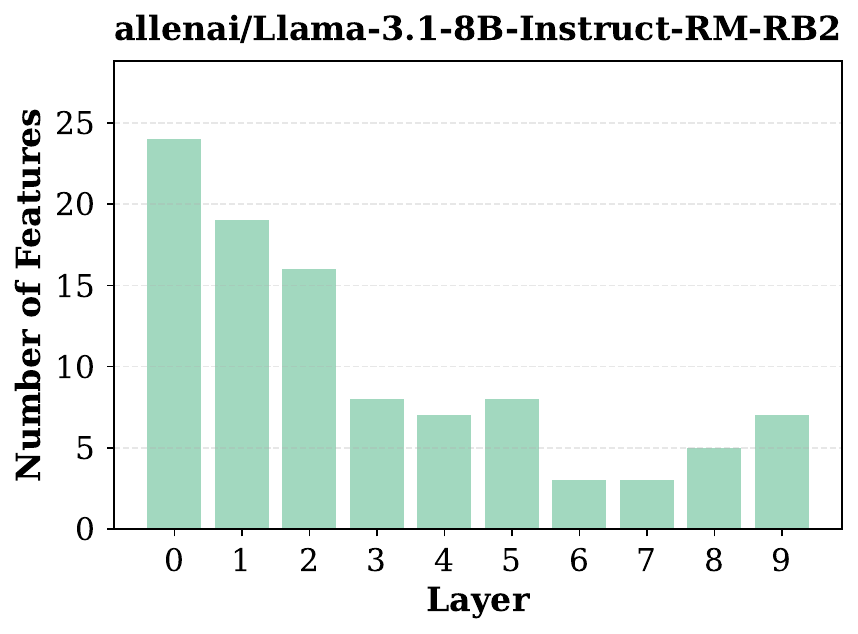}
        \caption{Llama-3.1-8B-Instruct}
        \label{fig:instruct}
    \end{subfigure}
    \hfill
    \begin{subfigure}[b]{0.48\columnwidth}
        \centering
        \includegraphics[width=\linewidth]{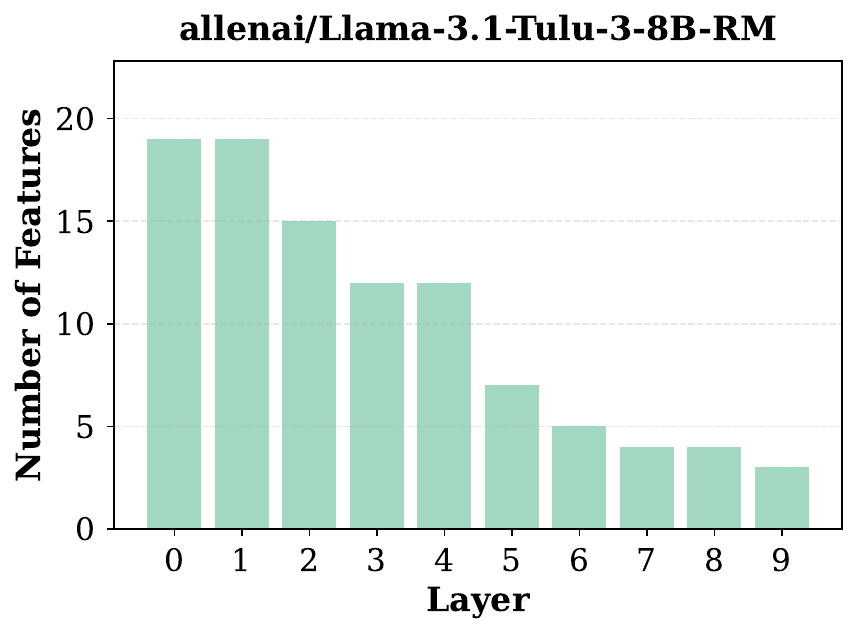}
        \caption{Llama-3.1-Tulu-3-8B}
        \label{fig:tulu}
    \end{subfigure}
    \caption{Layer-wise distribution of selected format-related features across four additional models.}
    \label{fig:format_features_dist_others}
\end{figure}

To further investigate whether the early-layer localization of format features is a general phenomenon, we extended our analysis to four additional models: LxzGordon/URM-LLaMa-3.1-8B, allenai/Llama-3.1-8B-Base-RM-RB2, allenai/Llama-3.1-8B-Instruct, and allenai/Llama-3.1-Tulu-3-8B. Following the same methodology as in the Section~\ref{sec:exp:layer}, we identified the top-100 format-related SAE features for each model based on their strength-stability scores.

As illustrated in Figure~\ref{fig:format_features_dist_others}, the layer-wise distribution of these features exhibits a consistent pattern across all evaluated models. The vast majority of format-sensitive features are concentrated in the early layers, particularly layers 0-3. This ubiquity suggests that formatting information is processed and encoded at the very beginning of the transformer computation, regardless of the initialization backbone used for reward modeling (e.g., whether initialized from a base or an instruct model). This finding reinforces our decision to focus interventions on these initial layers.

\subsection{Sensitivity to the Number of Probing Pairs}
\label{app:probe_size}

The main experiments use $N=500$ synthetic probing pairs. To test robustness to probe set size, we vary the number of synthetic pairs used for feature identification from 50 to 1000 on Skywork-Reward-Llama-3.1-8B. For each setting, we rerun the same strength-stability ranking, retain the top-10 features, and evaluate the resulting intervention on RM-Bench. We additionally report the overlap between each top-10 feature set and the default $N=500$ selection. The results are summarized in Table~\ref{tab:probe_size}.

\input{tables/probe_size}

The method is sensitive mainly in the extreme low-data regime. With only 50 probing pairs, feature identification becomes noticeably less stable, but once at least 100 pairs are available, both the selected features and downstream performance become comparable to the default configuration.

\subsection{Feature Identification from Existing Samples}
\label{app:existing_samples}

Although synthetic pairs provide a clean probe of formatting sensitivity, they are not essential to the method. We therefore repeat feature identification on Skywork-Reward-Llama-3.1-8B~\citep{liu2024skywork} using 500 randomly sampled Markdown/plain-text pairs from RM-Bench~\citep{liu2025rmbench}, matching the scale of the synthetic setup. The comparison between synthesized and existing-sample-derived features is shown in Table~\ref{tab:existing_samples}.

\input{tables/existing_samples}

The resulting features are highly consistent across data sources: 7 of the top-10 features identified from synthetic pairs also appear in the top-10 derived from RM-Bench. Steering with these existing-sample-derived features yields comparable overall gains and even stronger Hard-split improvement, indicating that the mechanism is not tied to synthesized probe data. Appendix~\ref{app:neuronpedia_features} further shows that the identified features activate on naturally occurring Markdown structures rather than dataset-specific artifacts.

\subsection{Top-K Selection Sensitivity Analysis}
\label{app:topk_ablation}

To understand the impact of the number of selected features on intervention effectiveness, we conduct a sensitivity analysis by varying the global top-$K$ value from 5 to 50. We evaluate each configuration on the Skywork-Reward-Llama-3.1-8B model using RM-Bench, reporting performance across all difficulty levels (Easy, Normal, Hard) and the overall average score.

\textbf{Experimental Setup.} We apply the same feature identification procedure described in Section~\ref{sec:method:identify} but vary the global top-$K$ selection parameter: $K \in \{5, 10, 20, 30, 50\}$. For each $K$ value, we identify the top-$K$ format-related features across layers 0-9, apply feature ablation interventions, and evaluate on RM-Bench. We report performance for each difficulty split (Easy, Normal, Hard) as well as the average score across all splits.

\begin{figure}[t]
  \centering
  \includegraphics[width=0.9\columnwidth]{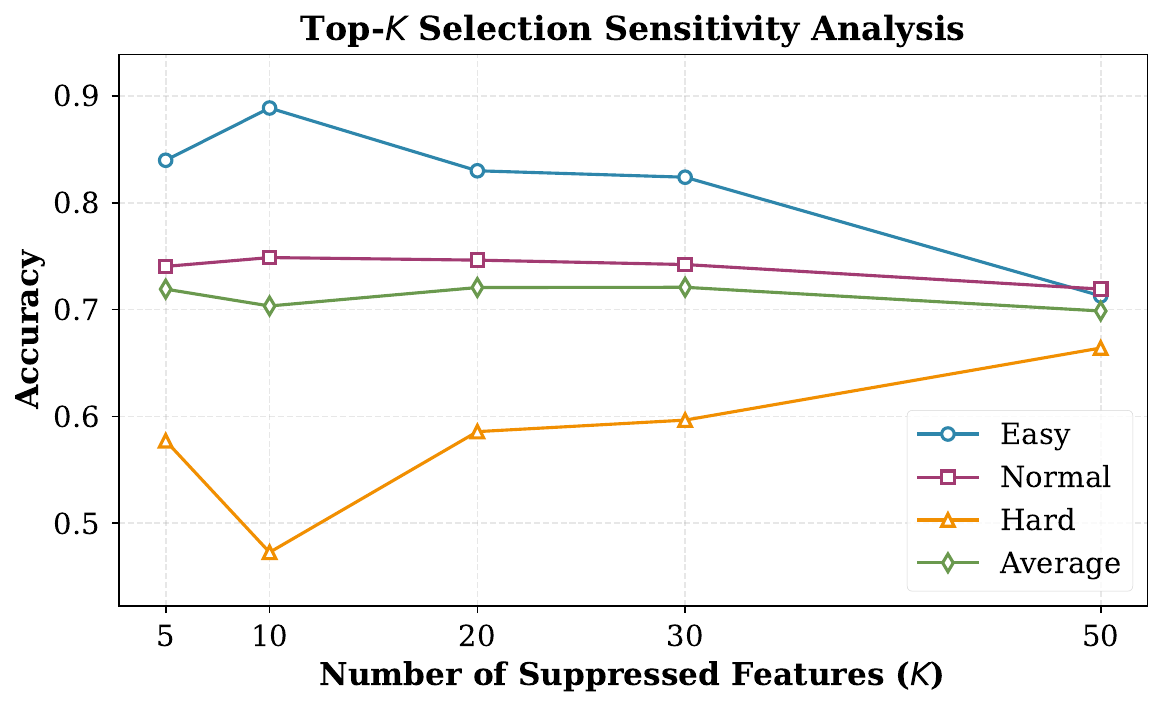}
  \caption{Performance sensitivity to the number of suppressed features ($K$). The plot shows accuracy across Easy, Normal, Hard difficulty levels and the overall average on RM-Bench.}
  \label{fig:topk_sensitivity}
\end{figure}

\textbf{Results.} Figure~\ref{fig:topk_sensitivity} shows the performance across different difficulty levels as a function of the number of suppressed features. A key observation is that as $K$ increases, the accuracy scores across Easy, Normal, and Hard difficulty levels converge toward each other, while Normal accuracy remains relatively stable (ranging from 0.740 to 0.749). This convergence pattern indicates that format-related feature suppression successfully eliminates format-based shortcuts in Easy tasks and format-induced challenges in Hard tasks, bringing all difficulty levels closer to the format-fair comparison represented by Normal difficulty. The average accuracy peaks at $K=30$, reaching approximately 0.721, after which it slightly declines as $K=50$ introduces less format-specific features.

This convergence demonstrates that SteerRM successfully suppresses superficial format-related signals, enabling reward models to focus on semantic content rather than surface-level formatting cues.

\subsection{Additional Format-Related SAE Features}
\label{app:neuronpedia_features}

In this section, we provide additional examples of format-related SAE features identified by our Strength-Stability criterion. These examples, spanning different Markdown elements such as lists, headers, and code blocks, further validate the effectiveness of our identification method. For each feature, we present its semantic interpretation (generated by GPT-4o-mini via Neuronpedia) and representative top-activating text snippets. We present four representative features below.

\begin{tcolorbox}[
    colback=cyan!2!white,
    colframe=cyan!25!gray,
    title={\textbf{Example: Feature 49738 (LLAMASCOPE-RES-131K Layer 1) -- Markdown headers and code blocks}},
    fonttitle=\bfseries\small,
    left=6pt,
    right=6pt,
    top=6pt,
    bottom=6pt,
    boxrule=0.5pt,
    breakable
]
\textbf{Semantic Interpretation from GPT-4o-mini:} terms related to software installation and usage instructions

\textbf{Top Activation Examples:}
\begin{itemize}
    \item ``ishing in30 seconds. Each test is run in a separate process. \colorbox{green!30}{\textasciigrave\textasciigrave\textasciigrave} \$ export PARALLEL\_SPLIT\_TEST\_PROCESSES=4\$ parallel''
    \item ``http://...).\colorbox{green!30}{\#\#} Links \colorbox{green!30}{\#\#} * [Requirements](http://...''
    \item ``
.com/...) base project.\colorbox{green!30}{\#\#} Installation \colorbox{green!30}{\#\#}See [installation on different platforms](http://...''
\end{itemize}

\emph{This feature activates strongly on structure-heavy segments typical of software documentation, such as Markdown headers (\texttt{\#\#}) and code blocks (\textasciigrave\textasciigrave\textasciigrave), illustrating the entanglement of formatting syntax with specific coding content.}
\end{tcolorbox}

\begin{tcolorbox}[
    colback=cyan!2!white,
    colframe=cyan!25!gray,
    title={\textbf{Example: Feature 41935 (LLAMASCOPE-RES-131K Layer 3) -- Markdown Emphasis Markers}},
    fonttitle=\bfseries\small,
    left=6pt,
    right=6pt,
    top=6pt,
    bottom=6pt,
    boxrule=0.5pt,
    breakable
]
\textbf{Semantic Interpretation from GPT-4o-mini:} references to data structures and functions used in programming, particularly in Python and related contexts

\textbf{Top Activation Examples:}
\begin{itemize}
    \item ``... reductions produce 1 row of output \colorbox{green!30}{\_}per group\colorbox{green!30}{\_}''
    \item ``... makes a \colorbox{green!30}{\_}big\colorbox{green!30}{\_} difference when you're mobile...''
    \item ``of user-defined functions for operations on expressions targeting the pandas backend: \colorbox{green!30}{**}element-wise\colorbox{green!30}{**}, \colorbox{green!30}{**}reduction\colorbox{green!30}{**}, and \colorbox{green!30}{**}analytic\colorbox{green!30}{**}...''
\end{itemize}

\emph{Although the semantic interpretation focuses on programming content, the activations reveal a clear sensitivity to Markdown emphasis markers, including both underscores (\texttt{\_}) for italics and asterisks (\texttt{**}) for bolding, often used to highlight technical terms.}
\end{tcolorbox}

\begin{tcolorbox}[
    colback=cyan!2!white,
    colframe=cyan!25!gray,
    title={\textbf{Example: Feature 46824 (LLAMASCOPE-RES-131K Layer 0) -- Universal Double-Asterisk Detector}},
    fonttitle=\bfseries\small,
    left=6pt,
    right=6pt,
    top=6pt,
    bottom=6pt,
    boxrule=0.5pt,
    breakable
]
\textbf{Semantic Interpretation from GPT-4o-mini:} formatting elements or emphasis within the text

\textbf{Top Activation Examples:}
\begin{itemize}
    \item ``... charcuterie. \colorbox{green!30}{**} Like pickles, mustard and sweet condiments...''
    \item ``... search results. \colorbox{green!30}{**} In fact, I have a `how to include history'...''
    \item ``person who invites those teachers out to teach workshops and then needs to promote them\colorbox{green!30}{**}  :D I think a lot of dance teachers are reluctant to encourage their students''
\end{itemize}

\emph{This layer-0 feature acts as a low-level detector for the \texttt{**} token sequence. It activates ubiquitously on double asterisks across diverse contexts, including Markdown emphasis and paragraph separation. This behavior illustrates that very early layers encode raw surface patterns prior to any contextual disambiguation.}
\end{tcolorbox}

\begin{tcolorbox}[
    colback=cyan!2!white,
    colframe=cyan!25!gray,
    title={\textbf{Example: Feature 20985 (LLAMASCOPE-RES-131K Layer 3) -- Code Comment Separators}},
    fonttitle=\bfseries\small,
    left=6pt,
    right=6pt,
    top=6pt,
    bottom=6pt,
    boxrule=0.5pt,
    breakable
]
\textbf{Semantic Interpretation from GPT-4o-mini:} programming or code syntax elements

\textbf{Top Activation Examples:}
\begin{itemize}
    \item ``\colorbox{green!30}{-} Call heater manager \colorbox{green!30}{* -} Call in activity manager * Call end stop manager \colorbox{green!30}{* -} Call LCD update */ void loop''
    \item ``R = delta radius \colorbox{green!30}{*} S = segments per second \colorbox{green!30}{*} A = Alpha ( Tower 1) diagonal rod trim \colorbox{green!30}{*}''
    \item ``\colorbox{green!30}{* -} The active serial input (usually USB) \colorbox{green!30}{* -} The SD card file being actively printed void get\_available\_commands () \{''
\end{itemize}

\emph{This feature detects comment separators (\texttt{*} and \texttt{-}) within multi-line code comments, illustrating how format-related patterns are intertwined with domain-specific content (code documentation).}
\end{tcolorbox}

\section{Computational Cost Analysis}

SteerRM is a training-free method, requiring no gradient updates or parameter optimization. The computational overhead primarily comes from SAE encoding/decoding during feature identification and inference-time interventions.

\textbf{Feature Identification Phase.} For each paired sample, we extract hidden representations and compute SAE latents across the selected layers (0-9). With 500 paired samples and 10 layers, this requires 5,000 forward passes through the SAE encoders and decoders. On our experimental setup (4× NVIDIA A800 80GB GPUs), this phase completes in a few minutes, with SAE models distributed across GPUs for parallel processing.

\textbf{Inference-Time Intervention.} During reward scoring, SteerRM intercepts hidden states at the layers where format-related features were identified, applies SAE encoding, ablates the selected features, and reconstructs the modified representations. The intervention overhead depends on the number of active layers (typically 3-5 layers in our experiments) and adds approximately 10-15\% computational cost compared to standard reward model inference, as SAE encoding and decoding are lightweight matrix operations.

Overall, the training-free nature of SteerRM eliminates the need for expensive model retraining or fine-tuning, while the inference-time overhead remains minimal. This makes it a practical and efficient approach for bias mitigation in reward models.

%% file: tables/probe_size.tex
\begin{table*}[t]
\centering
\small
\caption{
\textbf{Sensitivity to the number of synthetic probing pairs.}
Overlap is measured against the top-10 features identified with the default $N=500$ configuration. Numbers in parentheses show changes relative to the baseline model.
}
\label{tab:probe_size}
\setlength{\tabcolsep}{7pt}
\begin{tabular}{l c c c c c}
\toprule
\textbf{Synthetic Pairs ($N$)} & \textbf{Top-10 Overlap} & \textbf{Easy} & \textbf{Normal} & \textbf{Hard} & \textbf{Average} \\
\midrule
50 & 30\% & 80.8 (-8.1) & 70.3 (-4.6) & 48.1 (+0.8) & 66.4 (-3.9) \\
100 & 60\% & 88.0 (-0.9) & 75.1 (+0.2) & 48.7 (+1.4) & 70.6 (+0.3) \\
500 (default) & 100\% & 83.9 (-5.0) & 74.1 (-0.8) & 56.5 (+9.2) & 71.5 (+1.2) \\
1000 & 90\% & 84.2 (-4.7) & 74.0 (-0.9) & 54.1 (+6.8) & 70.8 (+0.5) \\
\bottomrule
\end{tabular}
\end{table*}

%% file: tables/existing_samples.tex
\begin{table*}[t]
\centering
\small
\caption{
\textbf{Feature identification from synthesized versus existing samples.}
Numbers in parentheses show changes relative to the baseline model.
}
\label{tab:existing_samples}
\setlength{\tabcolsep}{8pt}
\begin{tabular}{l l c c c c}
\toprule
\textbf{Method} & \textbf{Data Source} & \textbf{Easy} & \textbf{Normal} & \textbf{Hard} & \textbf{Average} \\
\midrule
Baseline & -- & 88.9 & 74.9 & 47.3 & 70.3 \\
SteerRM (Synth) & Synthesized Pairs & 83.9 (-5.0) & 74.1 (-0.8) & 56.5 (+9.2) & 71.5 (+1.2) \\
SteerRM (Existing) & Existing Samples & 82.0 (-6.9) & 74.4 (-0.5) & 60.5 (+13.2) & 72.3 (+2.0) \\
\bottomrule
\end{tabular}
\end{table*}